\documentclass{article} 
\usepackage{latex_template,times}
\usepackage{hyperref}
\usepackage{amsmath,amsfonts,amssymb}
\usepackage{algorithm}
\usepackage{algpseudocode}
\usepackage{url}
\usepackage{listings}
\usepackage{changepage}
\usepackage{float}
\usepackage{caption}
\usepackage{graphicx}
\usepackage{booktabs}
\usepackage{amsmath}
\usepackage{array}
\usepackage{appendix}
\usepackage{listings}
\usepackage{multirow}
\usepackage{subfigure}
\usepackage{arydshln}
\algnewcommand\INPUT{\item[\textbf{Input:}]}%
\algnewcommand\OUTPUT{\item[\textbf{Output:}]}%
\algnewcommand\STOP{\item[\textbf{Stop:}]}%
\newcolumntype{P}[1]{>{\centering\arraybackslash}p{#1}}

\newcommand{\BE}[1]{\textbf{#1 \quad}}

\title{Multimodal Multilabel Classification by CLIP}

\nipsfinalcopy 

\begin{document}
\author{
  Yanming Guo \\ 
  University of Sydney\thanks{This work partially fulfils the requirements for the Bachelor of Science (Honours) degree at the University of Sydney 2023. The GitHub repository is: \url{https://github.com/Yvnminc/Multimodal-Multilabel-Classification-by-CLIP}}
}

\maketitle

\begin{abstract}
Multimodal multilabel classification (MMC) is a challenging task that aims to design a learning algorithm to handle two data sources, the image and text, and learn a comprehensive semantic feature presentation across the modalities. In this task, we review the extensive number of state-of-the-art approaches in MMC and leverage a novel technique that utilises the Contrastive Language-Image Pre-training (CLIP) as the feature extractor and fine-tune the model by exploring different classification heads, fusion methods and loss functions. Finally, our best result achieved more than 90\% $F_1$ score in the public Kaggle competition leaderboard. This paper provides detailed descriptions of novel training methods and quantitative analysis through the experimental results.
\end{abstract}

\section{Introduction}
Multimodal Multi-label Classification (MMC) refers to a category of learning tasks that involve predicting one or more labels based on two modalities of information: images and text. MMC tasks present a greater complexity level than traditional single-modality, single-label classification tasks. This is primarily due to the requirement for models to process and interpret data from two distinct sources simultaneously. Designing an effective MMC system presents significant challenges, as it necessitates the creation of a learning structure capable of handling and fusing disparate input sources. This fusion must produce a comprehensive feature representation that integrates visual information with semantic content. The design of such a structure requires a deep understanding of both the visual and textual modalities. Furthermore, multi-label tasks inherently have more difficulty than single-label. In single-label classification tasks, the prediction with the highest posterior probability is typically selected as the output. Conversely, in multi-label classification tasks, the model is tasked with making predictions for each label, thereby increasing the complexity of the task. An additional layer of complexity in multi-label classification tasks arises from the intricate underlying dependencies that exist across labels. For instance, the label 'boat' is often associated with the label 'water'. Learning such relationships and incorporating them into the model further increases the complexity of the task. Therefore, the multimodal nature and the multi-label aspect of MMC tasks add to the complexity, making them a challenging yet intriguing area for machine learning research and development.

This study aims to classify 30000 multi-label images with textual descriptions of various sizes comprising 18 distinct classes. Each image is associated with one or more classes and a corresponding caption. The study has two primary objectives: Firstly, the study aims to experiment with different techniques that fuse the extracted features from the image and caption together. Secondly, the study plans to design and optimize the classification heads' structure and find a suitable design to enhance the models' performance. The performance of each technique and design choice will be compared and discussed, with the $F_1$ score serving as the primary performance metric. After obtaining the experiment's performance measures, the study will thoroughly analyze and justify the multimodal model's performance.

In today’s society, people consume vast amounts of multimedia content across various modalities, such as text, image, audio, video, and 3D. In many applications, utilising multiple modalities is extremely important, as it enhances performance by leveraging complementary information from different modalities, resulting in improved accuracy and robustness. It enables a richer representation of data by integrating distinct aspects conveyed by different modalities, facilitating more nuanced and accurate classification. Multimodal classification also helps handle ambiguity in classification problems by overcoming the limitations of individual modalities. The additional modalities essentially serve as extra opinions, effectively disambiguating data. Furthermore, studying the techniques and methods that deal with real-world data, which are usually multimodal, is crucial, enabling effective handling of diverse data types. Studies like this enable the development of models capable of cross-modal understanding and reasoning with practical implications in various domains. 

Learning a good feature presentation from a multimodel is a challenging task, as one must aim to learn robust and generalized features. Therefore, a straightforward solution is to train the model with many parameters on the sizable dataset. However, due to the limited training data, the increase in model parameters will lead to an overfitting problem. Moreover, the model size is constrained to 100 MB. Therefore, instead of training the model from scratch, we leverage the pre-trained model training on the data set and fine-tune it with the downstream dataset. We found that CLIP is the model that satisfies this objective well, as CLIP is trained on 400 million images and shows an impressive transferable ability. Therefore, our method is inspired by the motivation and fine-tuning of the classification heads. This work explores different classifiers, loss functions and fusion methods.

\section{Related Works}
\subsection{Multimodal Learning}
Multimodal learning, a burgeoning branch of machine learning, aims to amalgamate information from disparate modalities into a unified model structure. The input data could span across various formats, including but not limited to images, texts, audio, and video. Fusing this multimodal information, achieved by a learning algorithm, results in a comprehensive and precise understanding encapsulated within a unified latent feature space. In the realm of image-language learning, the input data generally consists of two components: the visual content (images) and the corresponding textual information that describes the visual content. This form of multimodal learning has a broad spectrum of real-world applications. A notable instance is Visual Question Answering (VQA), where algorithms are designed to interpret images and comprehend text, thereby enabling human-like interaction \cite{wu2017visual}. Other applications include text grounding, which aims to identify objects based on textual descriptions. \cite{Mun_2020_CVPR} proposed a method to localize segments within a video based on the classification sense. Multimodal learning holds immense potential and is a promising direction for future research.

\begin{figure}[t]
  \centering
  \includegraphics[width=\textwidth]{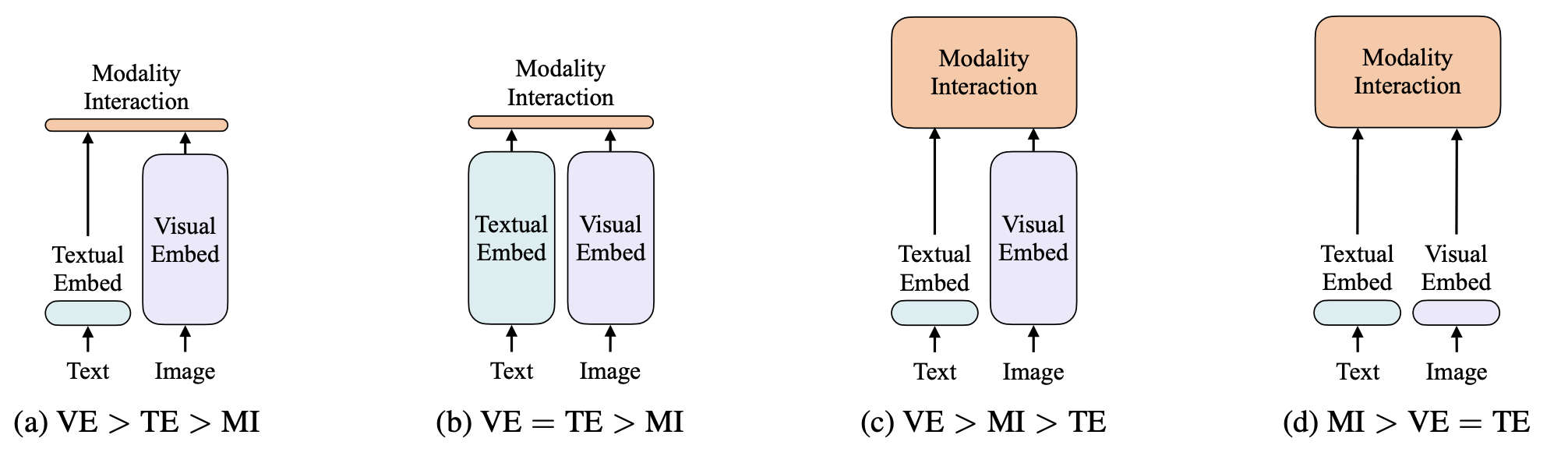}
  \caption{Four classes of the proposed methods according to the size of each module \cite{Kim2021ViLTVT}.}
  \label{fig:viltclass}
\end{figure}

The multimodal learning framework could be summarised as consisting of three parts: the text encoder, the image encoder and the modality fusion module. The major difference between each method could be the different backbone networks, fusion methods and the size of two encoders. The multimodel networks could be categorised into four classes \cite{Kim2021ViLTVT}, shown in Fig.(\ref{fig:viltclass}). Specifically, many models fall in the first category, exploiting a larger image encoder than the text encoder. For instance, the VSE++ \cite{faghri2018vse} is an improved visual semantic embedding (VSE) model architecture utilising the hinge-based triplet ranking loss to better compare the positive and negative samples. Moreover, SCAN \cite{lee2018stacked} leverages the cross-attention mechanism to understand better the highlighted part between the image and text feature.

As for the second class, where the text encoder and image encoder have a close computational size, Contrastive Language-Image Pre-training (CLIP \cite{radford2021learning}) is an example that leverages the contrastive learning paradigm to jointly train image-text pairs in a self-supervised fashion, guided by language-based supervision. The fundamental premise of CLIP is the shared semantic information between images and language. The model performance can be significantly enhanced by learning from images and text. Moreover, the availability of vast quantities of unlabelled images accompanied by captions facilitates the training process. CLIP has been trained on a dataset encompassing 400 million image-text pairs, thereby extending its transferability capabilities.

\begin{figure}[h!]
  \centering
  \includegraphics[width=\textwidth]{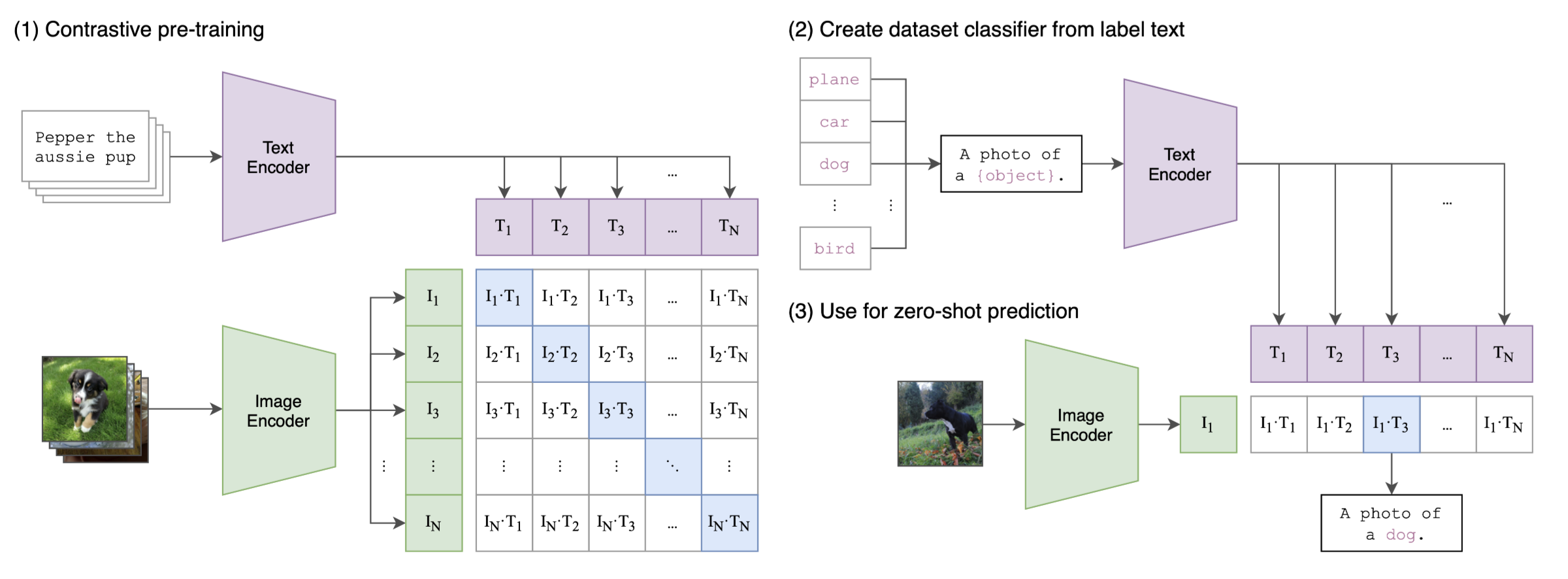}
  \caption{Structure of the CLIP network \cite{radford2021learning}.}
  \label{fig:clip}
\end{figure}

The architecture of CLIP is illustrated in Fig.(\ref{fig:clip}). It comprises two encoders, one for text and image, extracting features from the respective modalities. These features are subsequently fused through a joint multimodal embedding process, which involves computing the scaled pairwise cosine similarities after L2 normalization. The predicted embeddings are then contrasted with the positive samples, with the choice of positive and negative samples determined by their location on or off the diagonal, respectively (see Fig.(\ref{fig:clip})). This process encourages the model to maximize the similarity of matched image-text pairs while minimizing the similarity of unmatched pairs. This leads to a better understanding of the shared semantic space between images and text. In terms of applications, CLIP has demonstrated remarkable performance across various tasks, such as image classification and object detection, exhibiting its broad applicability and effectiveness in diverse scenarios.

Feature-wise Linear Modulation (FiLM \cite{perez2018film}) introduces a linear projection layer to transfer the text feature into a visual model. Therefore, FiLM could be classified into the third category with the higher computational cost for the fusion of two modality information.

\begin{figure}[t]
  \centering
  \includegraphics[width=\textwidth]{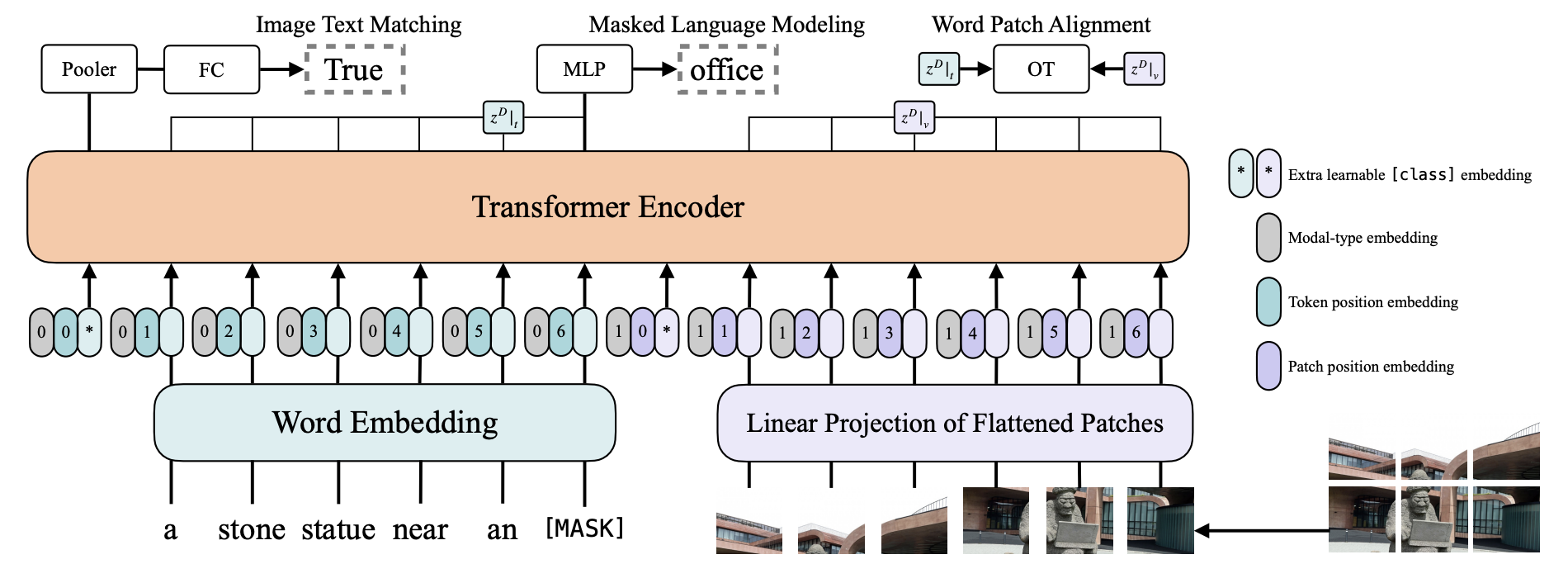}
  \caption{Structure of the ViLT network \cite{Kim2021ViLTVT}.}
  \label{fig:vilt}
\end{figure}

The last class where has a significant computational size of the modal interaction but less for the text and image encoder. Vision-and-Language Transformer (ViLT \cite{Kim2021ViLTVT}) is one example that is a new novel multimodel learning framework. The significant contribution of ViLT is simplifying and optimising the training speed for the multimodel learning framework. The input for the ViLT is the direct concatenation of the text and the image data. However, most of the computational cost is made for the fusion of two sources, which can be visually illustrated in Fig.(\ref{fig:vilt}).

\subsection{Backbones Networks}
In developing deep learning networks, there are three major backbones for multimodel learning: CNNs, RNNs and Transformers.

\BE{Convolutional Neural Networks (CNNs)}
CNNs are the conventional choice for the backbone for extracting the image feature in multimodel learning, which is optimised for the vision tasks. There are many existing CNN architectures. AlexNet \cite{Krizhevsky2012ImageNetCW} achieves impressive results in the ImageNet, which makes CNNs a popular choice for vision tasks. After that, VGGNet \cite{Simonyan2014VeryDC} and GoogleNet \cite{Szegedy2014GoingDW} were proposed and made CNNs deeper. However, with the depth increase, the gradient becomes smaller during the chain rule production. Therefore, ResNet \cite{He2015DeepRL} was postponed to address the problem. Another problem in deep neural networks is the high computational cost. Then, depthwise separable convolution \cite{Howard2017MobileNetsEC} is proposed to reduce the model parameters. Another solution to improve efficiency is utilising Neural Architecture Search (NAS) to find the network structure automatically. Therefore, EfficientNet \cite{Tan2019EfficientNetRM} compound scaling factor to find a high-performance network with less computational resource requirements.

\BE{Recurrent Neural Networks (RNNs)}
As for the text module, the conventional backbone network is RNNs \cite{medsker2001recurrent} to deal with sequential input such as texts or audio. RNNs could extract extended dependency features through sequential data. Long Short-Term Memory (LSTM \cite{graves2012long}) is a variant of RNNs that introduces the gates to balance the long and short dependency. Bidirectional Long Short-Term Memory (BiLSTM \cite{cornegruta2016modelling}) consists of two LSTMs to deal with the sequential input from left to right and right to left, respectively. However, the structure of LSTM is complicated both in implementation and computation. Therefore, a Gated Recurrent Unit (GRU \cite{dey2017gate}) is proposed consisting of two gates, and it achieves similar performance with LSTM but higher efficiency.

\BE{Transformer}
The Transformer \cite{vaswani2017attention} is the third neural network architecture after CNNs and RNNs. Compared with RNNs sequentially processing with text input. Transformers could deal with this parallelly by leveraging the attention mechanism. Bidirectional Encoder Representations from Transformers (BERT \cite{Devlin2019BERTPO}) is a thriving network structure in Natural Language Processing through a directional encoder to better learn the context information. Opposite to BERT, GPT \cite{Radford2018ImprovingLU, Radford2019LanguageMA, Brown2020LanguageMA} utilizes a decoder for the text generation task. The parameters of GPT3 increase to 175 billion, proving the Transformer's scalability. Moreover, the Transformer becomes a uniform architecture for vision and text modules. Vision Transformer \cite{Dosovitskiy2020AnII} is proposed to show that self-attention could also benefit vision tasks. To better improve the performance of ViT, SWin Transformer \cite{Liu2021SwinTH} was proposed with a CNN-like hierarchical structure.

\section{Method}
\subsection{Feature Extraction}
Given the dataset contains both image and textual information, it is natural to ask how much content we should leverage for the multilabel classification task. Intuitively, natural languages, especially image captions, can better supervise image classification tasks because they usually contain the concepts of images. Therefore, we decided to use both of them in our project. Specifically, we harness the power of the CLIP model, which learns patterns from supervision contained in language information \cite{radford2021learning}.

The philosophy of CLIP can be explained into two downstream tasks: image feature extraction and text feature extraction. 

\BE{Image Feature Extraction} For image feature extraction, a Vision Transformer (ViT) \cite{Dosovitskiy2020AnII} is used. Initially, the input image is partitioned into equal-sized patches. These patches are then flattened for ease of processing. However, given the propensity for the dimensionality to become excessively large post-flattening, linear mapping is employed to reduce the data to a more manageable, lower-dimensional space. Subsequently, positional embeddings are appended to these patches, which allows the model to effectively keep track of the location of each patch within the original image. As a fourth step, a class token is added, which aids in deriving the encoded characteristics of the input image. This data is then supplied to the encoder for further refinement and extraction.

\BE{Text Feature Extraction} On the other hand, a language Transformer \cite{vaswani2017attention} is used to extract text features. Specifically, as a sequence of text enters into an encoder, it first traverses a self-attention layer that allows the encoder to evaluate the rest of the words in the input sequence during the word encoding process. A fully connected feed-forward neural network then processes the product of this self-attention operation. Although the quantity of parameters in each encoder's feed-forward neural network is identical, their functions operate independently. The decoder component mirrors this layered structure but incorporates an additional Attention layer. This layer aids the decoder in focusing on the word that corresponds to the input sentence, further enhancing the feature extraction process.

\begin{figure}[h!]
  \centering
  \includegraphics[width=\textwidth]{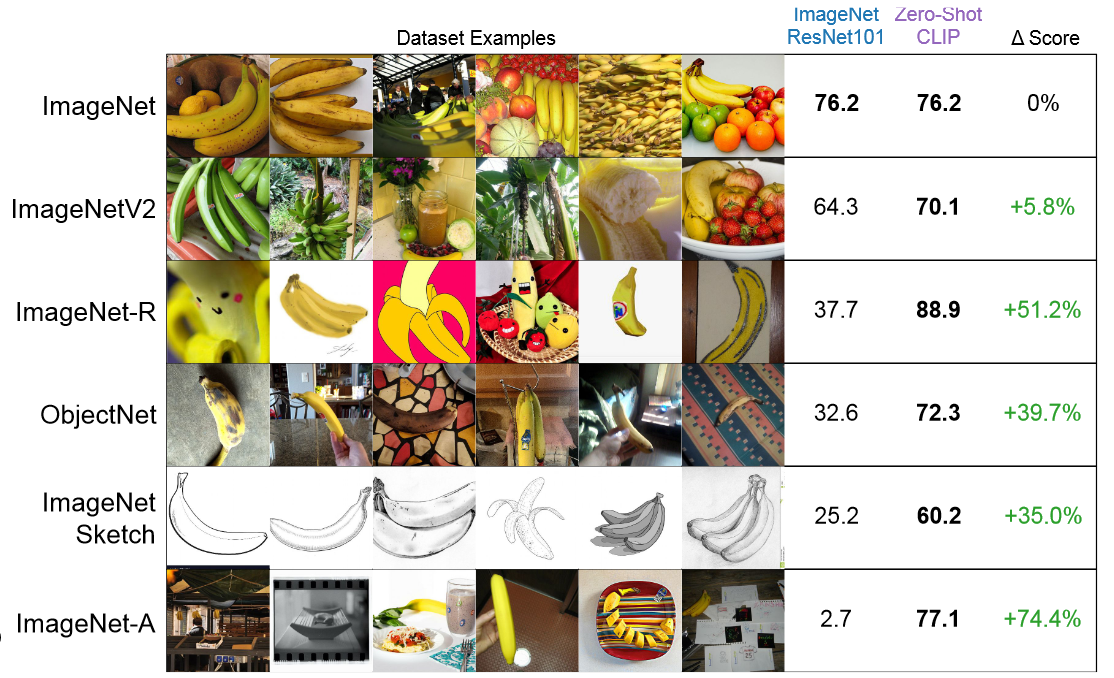}
  \caption{Generalization ability of ResNet-101 vs. Zero-Shot CLIP \cite{radford2021learning}.}
  \label{fig:clip_feature}
\end{figure}

By contrastively combining these two processes during the training, CLIP can learn a shared semantic space between images and text. According to \cite{radford2021learning}, CLIP is much more robust to distribution shift than standard ImageNet models such as ResNet families \cite{he2015deep}. As shown in Fig.(\ref{fig:clip_feature}), Zero-Shot CLIP demonstrates its dominated generalization ability to unseen data. This supports that CLIP's capability to learn joint representations for images and text from the internet-scale dataset significantly improves the quality of the extracted features, thereby enhancing the accuracy of our task.

\subsection{Fusion}
Since it is a multi-modality task, we aim to find a way to combine the knowledge from each modality to enhance the model performance. Fusion is the answer. Fusion can be implemented at different levels, such as feature and decision levels, depending on where the fusion occurs. In this project, we implement three different fusion methods: one feature-level fusion (i.e., concatenation fusion), one decision-level fusion (i.e., sum fusion) and one mixed fusion.

\BE{Concatenation Fusion} Concatenation fusion combines image and text features before feeding into the classification head. It first extracts the features of both the images and captions as two matrices with the same dimension. Then, merge the text and image information by flattening them horizontally, resulting in a combined matrix with twice as many dimensions. This simple technique can guide the model to learn more accurate feature data than relying on text or image features alone.

\BE{Sum Fusion} Similar to concatenation fusion, the philosophy of sum fusion is addition. However, different to concatenation fusion, instead of fusing features, sum fusion adds the output of image features and text features after being fed into the classification head. As a result, this approach will treat the output of the two modalities as an input to the meta-classifier. That is to say. The final decision comprehensively considers the results of both image and text features.

\BE{Mixed Fusion} Mixed fusion combines the idea of concatenation fusion and sum fusion. The procedure begins with concatenation fusion to merge the image and text features. Then, train the concatenated attributes to obtain the output. Next, feed the image features and text features separately into the classification head to obtain the second and the third outputs. Finally, sum fusion is used to sum up three output features and retrain them with a meta-classifier to produce the final outcome. 

Using fusion methods has several potential benefits in general. Firstly, fusion helps improve the robustness of a model because other modalities can compensate for the unreliability of one modality. Secondly, fusion provides a more comprehensive understanding of the sentiment being expressed because different data sources can provide different perspectives on a problem. For example, the image data could provide information about what the objects look like, and the text data could provide information about the image's content. Furthermore, fusion methods often have more excellent generalization capabilities, especially when they are designed to handle input from multiple modalities. This can lead to better performance on unseen data.

Specifically, mixed fusion is a novel fusion approach with multiple advantages. Firstly, combining feature fusion and decision fusion leverages the strengths of both methods, potentially leading to more accurate and robust predictions. Secondly, processing image and caption features both together and separately ensures a comprehensive understanding of the data. This could be particularly beneficial in tasks where the images and captions provide complementary information.
 
\subsection{Loss Function}
 The loss function is a fundamental aspect of deep learning that evaluates the difference between the predicted and actual output and is employed to update the model parameters during training. The choice of loss function is influenced by the nature of the task and the data being analyzed. For instance, the cross-entropy loss function is commonly employed in binary classification tasks, while the mean squared error (MSE) or mean absolute error (MAE) loss functions are prevalent in regression tasks. Recently, a rising interest has been in designing new loss functions that embed domain-specific expertise, such as domain adaptation and resilience to adversarial attacks. Selecting and creating a loss function is a crucial factor in developing deep learning models, as it directly influences their performance and generalization capacity.

In developing our model, we experimented with three different types of loss functions: binary cross-entropy, focal loss, and asymmetric loss.

\BE{Binary Cross Entropy}
Binary cross-entropy is a common loss function in deep learning for multilabel classification tasks. In the context of multi-label classification, it is a common practice to simplify the problem by transforming it into a sequence of binary classification tasks. Specifically, in this approach, the base network generates a single logit, $z_k$, for each of the $K$ labels. The activation of each logit is carried out independently, using a sigmoid function $\delta(z_k)$. The ground truth for class $k$ is denoted as $y_k$. The overall classification loss, $L_{tot}$, is derived by summing up the binary loss for all $K$ labels:
\begin{equation} \label{bin_loss}
L_{tot} = \sum_{k = 1}^K L(\delta(z_k), y_k),
\end{equation}
where the binary loss per label, L, is given by:
\begin{equation} \label{bin_loss_per_label}
L= -yL_+ - (1-y)L_-,
\end{equation}
where y is the ground-truth label, and $L_+$ and $L_-$ are the positive and negative loss parts, respectively.
Binary cross-entropy loss has several advantages: it is differentiable, computationally efficient, and easy to interpret.

\BE{Focal Loss}
The Focal loss function is a specialized loss function tailored for multilabel classification problems characterized by imbalanced class distributions \cite{lin2017focal}. In such scenarios, the distribution of samples may be skewed towards one or a few dominant classes, whereas the minority classes may have an insufficient number of instances. As a result, the model's predictions may be biased toward the dominant classes, thereby leading to suboptimal performance on the minority classes. Focal loss addresses this issue by selectively down-weighting the contribution of the easy examples (i.e., well-classified negative samples) and emphasizing the contribution of the harder examples (i.e., misclassified rare positive samples) during training. Specifically, the Focal loss function introduces a modulating factor that magnifies the impact of the loss incurred by the hard-to-classify samples and diminishes the impact of the loss incurred by the easy-to-classify samples. This modulating factor is governed by a user-specified parameter known as the focal parameter, which governs the degree of emphasis on the hard examples.
Focal loss is obtained by further setting $L_+$ and $L_-$ as:
\begin{equation} 
\label{eq: focal}
\begin{cases} 
      L_+ = (1-p)^\gamma log(p)\\
      L_- = p^\gamma log(1-p)\\
   \end{cases}
   ,
\end{equation}
where $p = \delta(z)$ is the network’s output probability and $\gamma$ is the focusing parameter ($\gamma = 0$ yields binary cross-entropy). By setting $\gamma > 0$ in Eq.(\ref{eq: focal}), the contribution of easy negatives (having low probability, $p << 0.5$) can be down-weighted in the loss function, enabling to focus more on harder samples during training.
Focal loss has been shown to improve the performance of deep learning models on imbalanced multilabel classification tasks and has become a popular choice in the research community.

\BE{Asymmetric Loss}
In multi-label classification, there is often an imbalance between the number of positive and negative labels due to the large number of possible labels. Typically, an image will have only a few positive labels and many negative ones. This imbalance can dominate the optimization process during training and lead to insufficient emphasis on gradients from positive labels, resulting in lower accuracy. To address this issue, a new asymmetric loss function called ASL has been introduced \cite{ben2020asymmetric}. ASL operates differently on positive and negative samples, allowing for dynamic down-weighting and hard-thresholding of easy negative samples and the removal of possibly mislabelled samples. ASL has two main features: First, the positive and negative samples are modulated separately using different exponential decay factors to focus on hard negatives while maintaining the contribution of positive samples. Second, the probabilities of negative samples are shifted to discard very easy negatives using hard thresholding, enabling the removal of challenging negative samples suspected to be mislabelled. 

To define Asymmetric Loss (ASL), the two mechanisms, asymmetric focusing and probability shifting, are integrated into a unified formula: 
\begin{equation} \label{ASL}
\begin{cases} 
      L_+ = (1-p)^\gamma log(p)\\
      L_- = (p_m)^\gamma log(1-p_m)\\
   \end{cases}
   ,
\end{equation}
where the shifted probability $p_m$ is defined as:
\begin{equation} \label{pm}
p_m = \max(p-m,0).
\end{equation}
ASL allows us to apply two types of asymmetries for reducing the contribution of easy negative samples to the loss function: soft thresholding via the focusing parameters $\gamma_- > \gamma_+$ and hard thresholding via the probability m.

Experimenting with different types of loss functions is essential when developing deep learning models because the choice of loss function significantly impacts the model's performance. Different loss functions are designed to optimize for different objectives, such as classification accuracy or robustness to outliers. By trying out different loss functions, we can identify which loss function is best suited to the task at hand and can fine-tune the model to achieve better performance. Besides the common choice of binary cross-entropy and focal loss, we implemented the novel asymmetric loss (ASL), which addresses the inherent imbalance nature of multi-label datasets.

\subsection{Exponential Moving Average (EMA)}
The exponential moving average (EMA) represents a technique utilized in neural network training that may enhance model accuracy. Specifically, rather than adopting the optimized parameters derived from the final training iteration as the ultimate parameters for the model, the EMA of the parameters obtained throughout all the training iterations is employed. In supervised learning, a neural network is trained on a labelled dataset $D$ consisting of $N$ instances to optimize the network parameters $\Theta$ by minimizing the associated loss function $L(D; \Theta)$. When the neural network does not overfit the training dataset, and the validation dataset is representative of the training dataset, the parameters that minimize the training loss also minimize the validation loss. These parameters are called $\Theta^*$, corresponding to a local optimum.

However, optimizing the parameters using the entire training dataset can be computationally expensive, particularly with standard optimization techniques such as gradient descent, given the large size of $N$. Therefore, training is often performed on small-sized samples randomly extracted from the training dataset, referred to as mini-batches. Although mini-batches introduce noise into the training process, this noisy gradient may sometimes be advantageous to the optimization process, resulting in a better local optimum than the one obtained with the entire dataset. Nonetheless, as parameter optimization approaches convergence, the optimized parameters may not represent a local minimum. Still, they may fluctuate around it, representing a disadvantage of the training noise introduced by mini-batches.

To mitigate this problem, smoothing techniques, such as the exponential moving average (EMA), can be employed to reduce the optimized parameter fluctuation noise. The parameter EMAs are more likely to be closer to a local minimum. Specifically, the optimized parameters after each update step are $\Theta_1, \Theta_2,...,\Theta_t, ...,\Theta_n$, forming a sequence of time-series data that contains noise. EMA can be applied to this sequence to improve model generalization.

The principle behind EMA is to smooth out the noise in the training process and reduce the variance of the model. To remove the noise, the EMA for the optimized parameter is computed as:

\begin{equation} \label{ema_eq}
EMA_t = \begin{cases} 
      \Theta_1 & t = 1 \\
      \alpha EMA_{t-1}+(1-\alpha)\Theta_t & else \\
   \end{cases}
   .
\end{equation}

As shown in Eq.(\ref{ema_eq}), EMA is implemented by taking a weighted average of the current and previous model parameters. The weight given to the current parameters is higher than the previous ones. This ensures that the current parameters have a larger influence on the new parameter values while still retaining some of the information from the previous parameters. The weight given to the current parameters is controlled by a hyperparameter $\alpha$ called the decay rate. The higher the decay rate, the more weight is given to the current parameters and the more the model is adapted to the current data.

EMA is typically used in the context of stochastic gradient descent optimization, where the model parameters are updated after each mini-batch of data. Using EMA, the model can adapt to the current mini-batch of data while retaining some information from the previous mini-batches, resulting in more stable and better-performing models.

The exponential moving average (EMA) provides several benefits when used in deep learning models. First, it stabilizes the training process by reducing the variance of the gradients during training. This helps to smooth out fluctuations in the gradient updates and provides a more stable estimate of the gradient. Second, EMA improves generalization performance by preventing overfitting and reducing the impact of outliers and noise in the training data. Third, it increases the model's robustness to changes in the training data, preventing overreaction to small changes. Finally, EMA can improve the convergence speed of the training process by reducing oscillations in the loss function and helping the model to reach the optimal solution faster.

The developmental trajectory of Model EMA was initially paved with Snapshot Ensembles \cite{huang2017snapshot}. This approach entailed utilizing a cyclic learning rate to obtain multiple model checkpoints from a single training session, followed by the ensemble averaging of predictions. This technique represented a favourable alternative to the erstwhile norm of training multiple models from scratch to create the ensemble. Subsequently, the logical progression from prediction averaging was to incorporate weight averaging, leading to the advent of Stochastic Weight Averaging (SWE) \cite{izmailov2018averaging}. Before SWE, Mean Teacher employed a running average of model weights \cite{tarvainen2017mean}. Today, EMA remains a prevalent tool in recent semi-supervised and unsupervised methodologies, such as BYOL \cite{grill2020bootstrap}.

\subsection{Classification Head}
Given that our task is image classification, the goal is to map the high-level features learned during the feature extraction phase to the classes the model is trying to predict. Since the feature extraction and classification process is separated into two parts, we have to design a classification head to make final predictions. Specifically, the classification head is usually a more straightforward structure, often consisting of several fully connected layers. In this project, we explore two types of classification heads, the vanilla Multi-Layer Perceptron (MLP) and Gated Multi-Layer Perceptron (gMLP) \cite{liu2021pay}.

\BE{Multi-Layer Perceptron (MLP)}
MLP is a fully connected neural network consisting of three layers: an input layer, a hidden layer, and an output layer. Intuitively, the input layer receives the data, the hidden layers process the data, and the output layer produces the final result.

The learning process in MLPs is achieved through backpropagation. During the backpropagation, the model predicts based on the input data. The output is compared to the desired output (i.e., the target), and the difference (i.e., the error) is calculated. This error is then propagated back through the network, adjusting the weights of the connections between the neurons in the process. This process guides the model to learn the underlying structure of the data.

\begin{figure}[h!]
  \centering
  \includegraphics[width=0.7\textwidth]{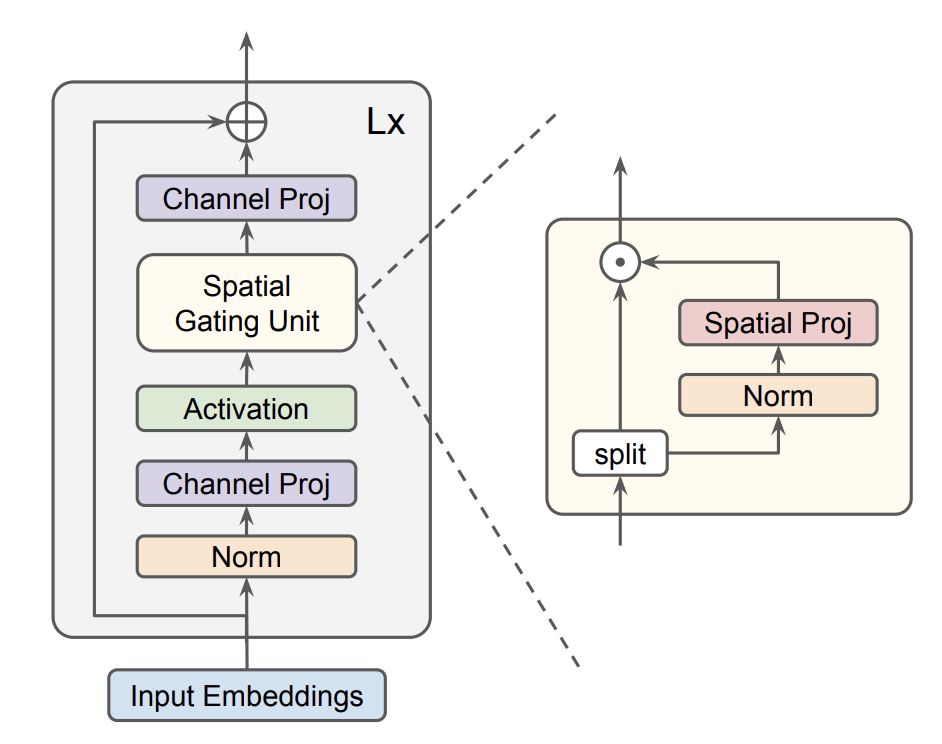}
  \caption{Strucuture of gMLP \cite{liu2021pay}.}
  \label{fig:gmlp}
\end{figure}

\BE{Gated Multi-Layer Perceptron (gMLP)}
gMLP is an improved version of MLP. The basic principle of gMLP is the same as that of MLP. The difference is that gMLP incorporates the attention mechanism into the original structure. The main idea of gMLP is leveraging the gate to decide the block or accretion of the information. The gate mechanism makes gMLP capable of dealing with sequential data by extracting long-dependent information, as illustrated in Fig.(\ref{fig:gmlp}).

A standard MLP layer is mathematically defined as $f_{W,b}(Z)$, where $W$ is the weight and $b$ is the bias term. $Z$ is input as follows:
\begin{equation}
f_{W,b}(Z) = WZ + b.
\end{equation}
The gate $s(\cdot)$ is defined in Eq.(\ref{eq:gate}):
\begin{equation}
s(Z) = Z \odot f_{W,b}(Z).
\label{eq:gate}
\end{equation}
Therefore, as stated in Fig.(\ref{fig:gmlp}), the input $Z$ is divided into two independent parts, $Z_1$ and $Z_2$, in the channel dimension. Then the gate operation is illustrated in Eq.(\ref{eq:gatefunction}):
\begin{equation}
s(Z) = Z_1 \odot f_{W,b}(Z_2).
\label{eq:gatefunction}
\end{equation}

The standard MLPs could learn non-linear features by utilising the activation functions \cite{nair2010rectified, xu2015empirical, ramachandran2017searching, hendrycks2016gaussian, he2015delving} since the CLIP already extracts the feature representation with high transferability. Therefore, training an MLP classification should be lightweight and not very complicated. By incorporating the attention mechanism into the MLP structure, gMLP allows the model to focus selectively on parts of the input sequence that are more relevant to a given task. This can be particularly beneficial for tasks involving long sequences where vital information can be located anywhere.

From another perspective, using the classification heads alone is advantageous. Specifically, separating feature extraction and classification allows for modularity and flexibility in designing machine learning models. For example, it's possible to use the exact feature extraction mechanism with different classification heads, depending on the specific task at hand. Conversely, it's also possible to experiment with different feature extraction mechanisms while keeping the same classification head.

\newpage
\section{Experiments and Results}
We aim to design our experiments into three main parts. The first part aims to choose the best pipeline of our methods by parallelly comparing different techniques such as classification heads, fusion methods and loss functions. Then the second part is to fine-tune the best pipeline by modifying hyperparameters such as the layer structure of the classification heads, learning rates,  epoch numbers, etc. Finally, there will be an ablation study to investigate the effect of a specific parameter.

\subsection{Experiment Setups}
\subsubsection{Evaluation Metric}
This project uses the mean $F_1$ score as the evaluation metric. The $F_1$ score is a widely accepted measure in the field of information retrieval, which calculates accuracy based on the statistical values of precision and recall. Precision refers to the proportion of true positives (i.e., $tp$) out of all predicted positives (i.e., $tp + fp$). In contrast, recall refers to the proportion of true positives (i.e., $tp$) compared to all actual positives (i.e., $tp + fn$). The $F_1$ score can be formulated as:
\begin{equation}
    F_1 = 2\frac{\text{Precision} \cdot \text{Recall}}{\text{Precision + Recall}},~\text{where}~\text{Precision} = \frac{tp}{tp + fp},~\text{Recall} = \frac{tp}{tp + fn}.
\end{equation}
\subsubsection{Hyperparameter Initialization}
For fair comparisons, we set the training settings to be the same for all the methods. We summarized our experiment setups in Tab.(\ref{tab: experiment setups}). Also, we split the given data into 25,000 training data and 5,000 validation data and used a training batch size equivalent to the training data size. Note that in Sec.(\ref{Sec: hyperparameter}), we will tune some of these hyperparameters to achieve the best $F_1$ score.

\begin{table}[!h]
    \centering
    \resizebox{\linewidth}{!}{
    \begin{tabular}{c|c|c|c|c|c}
    \hline
     \textbf{Learning Rate} & \textbf{Activation Function} & \textbf{Optimizer} & \textbf{Dropout Rate} & \textbf{Epoch Number} \\
    \hline
    \centering 0.001 & GeLU & Adam & 0.5 & 200\\
    \hline
    \end{tabular}}
    \caption{Experiment Setups.}
    \label{tab: experiment setups}
\end{table}

\subsection{Comparison Methods}

\subsubsection{Classification Head}
\begin{table}[!h]
    \centering
    \scriptsize
    \resizebox{0.8\linewidth}{!}{
    \begin{tabular}{c|c|c}
        \hline
        \multicolumn{1}{c|}{\textbf{Classification Head}} & 
	\multicolumn{1}{c|}{\textbf{Layer Number}} & 
		\multicolumn{1}{c}{\textbf{Validation $F_1$ score (\%)}} \\
		\hline
        MLP & 3 & \textbf{85.533 $\pm$ 0.001}\\
        gMLP & 3 & 83.506 $\pm$ 0.000\\
        \hline
	\end{tabular}}
	\caption{Comparison of the validation $F_1$ score of different classification heads for 200 epochs. We use \textbf{bold} to represent the best result.}
	\label{tab:head}
\end{table}

We compare the performance of two classification heads (i.e., MLP and gMLP) and report the averaged $F_1$ score with standard deviations in Tab.(\ref{tab:head}). Both models have the same layer structure and were trained over 200 epochs with the same training settings. For simplicity, we use the image features only. The MLP had achieved a $F_1$ score of 85.533\% on the validation set. In contrast, the gMLP head achieved a slightly lower $F_1$ score of 83.506\%, representing MLP outperforms gMLP with an improvement of 2\%. The result seems counterintuitive to the theoretical analysis of gMLP as it is an improved version of MLP that incorporates the attention mechanism.

\begin{figure}[!h]
  \centering
  \includegraphics[width=0.7\textwidth]{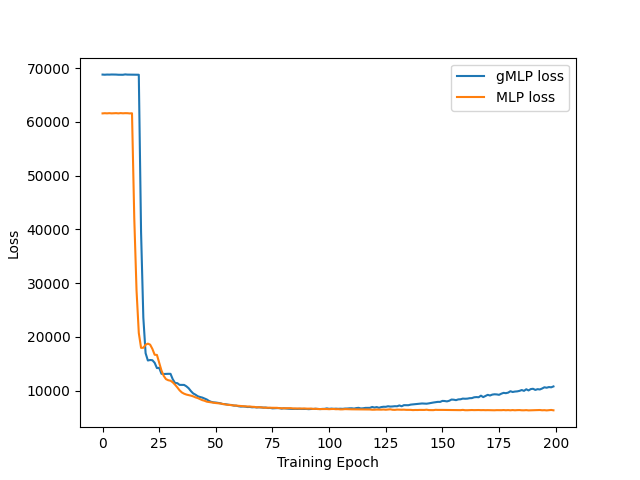}
  \caption{Validation loss for MLP and gMLP.}
  \label{fig:mlploss}
\end{figure}

To further validate the performance of MLP and gMLP, we plot the line graph for the validation loss of these two methods (see Fig.(\ref{fig:mlploss}). It could be found that initially, the validation loss of the gMLP is higher. This is probably because the gMLP is more complex than MLP. At around the 100th epoch, they can achieve almost the same performance. However, possibly due to the complexity of gMLP, the loss begins to increase after around 125 epochs, indicating an overfitting problem occurs for gMLP. In contrast, the loss drops faster for MLP with a more stable training process afterwards. This result suggests that a more complex model structure may not perform better in every case. \textbf{Therefore, we choose to use the MLP as the classification head for future experiments.}

\subsubsection{Fusion}
\begin{table}[!h]
    \centering
    \scriptsize
    \resizebox{0.6 \linewidth}{!}{
    \begin{tabular}{c|c}
        \hline
	\multicolumn{1}{c|}{\textbf{Fusion Methods}} & 
		\multicolumn{1}{c}{\textbf{Validation $F_1$ score (\%)}} \\
		\hline
        Image Features Only & 85.375 $\pm$ 0.051\\
        Text Features Only & 82.493 $\pm$ 0.000 \\
        Sum Fusion & \textbf{86.338 $\pm$ 0.005}  \\
        Concatenation Fusion & 85.779 $\pm$ 0.000 \\
        Mixed Fusion & 86.120 $\pm$ 0.011 \\
			\hline
	\end{tabular}}
	\caption{Comparison of the validation $F_1$ score of different fusion methods. We use \textbf{bold} to represent the best result.}
	\label{tab: fusion}
\end{table}
\begin{figure}[h]
  \centering
  \begin{minipage}[b]{0.49\textwidth}
    \includegraphics[width=\textwidth]{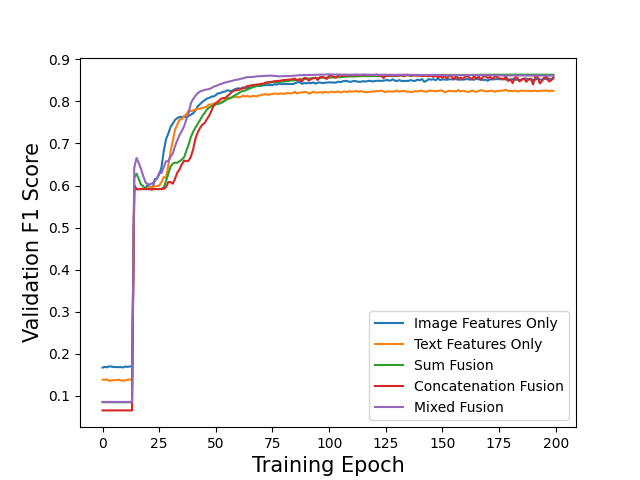}
    \caption{An overall look at validation $F_1$ score of different fusion methods}
    \label{fig: $F_1$}
  \end{minipage}
  \hfill
  \begin{minipage}[b]{0.49\textwidth}
    \includegraphics[width=\textwidth]{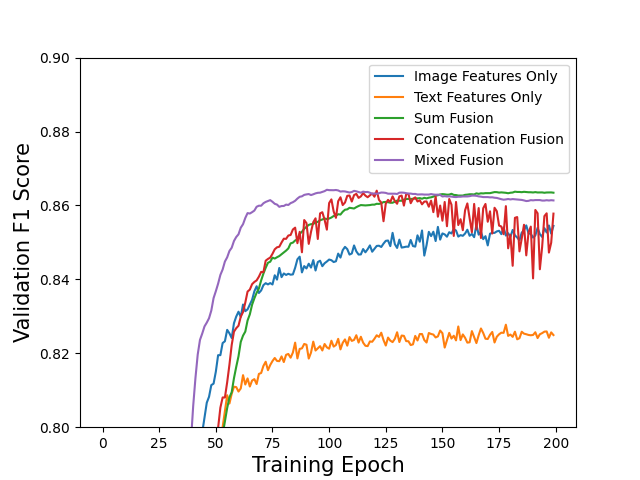}
    \caption{A closer look at validation $F_1$ score of different fusion methods}
    \label{fig: closer_$F_1$}
  \end{minipage}
\end{figure}

We compare the performance of different fusion methods to the baseline models that use only image features or text features. The averaged $F_1$ scores with standard deviations are summarized in Tab.(\ref{tab: fusion}). Note that we report the last $F_1$ scores for a fair comparison instead of the best $F_1$ scores. To reflect the actual effectiveness of each fusion method, we control all the hyperparameters in the same way, which includes hyperparameters of the training procedure and the structure of the classification head. As illustrated by Tab.(\ref{tab: fusion}), fusion methods outperform the baseline, supporting our previous statements that fusion leads to better performance in multi-modality tasks. Among all the fusion methods, sum fusion achieves the best performance with a low variance, followed by mixed fusion. 

Theoretically, mixed fusion should perform best since it combines concatenation and sum fusion. Empirically, by plotting the change in validation $F_1$ scores against training epochs (see Fig.(\ref{fig: $F_1$}) and Fig.(\ref{fig: closer_$F_1$})), we find that mixed fusion outperforms other methods in the first 100 epochs, but begins to decrease afterwards. This is probably attributed to the fluctuations in concatenation fusion (i.e., as part of the mixed fusion), where it appears to be unstable after around the 150th epoch. Sum fusion, on the other hand, still has a trend to further improve the validation $F_1$ score after the 200th epoch. \textbf{Therefore, we choose to apply sum fusion for future experiments} since we believe it will further boost the performance if we carefully tune the training epochs and other hyperparameters.

\subsubsection{Loss Functions}
\begin{table}[!h]
    \centering
    \scriptsize
    \resizebox{0.6 \linewidth}{!}{
    \begin{tabular}{c|c}
        \hline
	\multicolumn{1}{c|}{\textbf{Loss Functions}} & 
		\multicolumn{1}{c}{\textbf{Validation $F_1$ score (\%)}} \\
		\hline
        Binary Cross-Entropy&\textbf{86.443 $\pm$ 0.005}\\
        Focal Loss & 86.432 $\pm$ 0.005 \\
        Asymmetric Loss & 85.051 $\pm$ 0.005  \\
        \hline
	\end{tabular}}
	\caption{Comparison of the validation $F_1$ score of different loss fuction. We use \textbf{bold} to represent the best result.}
	\label{tab: loss}
\end{table}

We compared the performance of three different loss parameter settings. The first setting sets the negative gamma  = 0, positive gamma = 0, and clipping  = 0. This is the equivalent of essential Binary Cross-Entropy Loss. The second setting sets the negative gamma  = 3, positive gamma = 3, and clipping  = 0. This is the equivalent of focal loss with a focusing parameter of 3. The third setting sets the negative gamma  = 4, positive gamma = 1, and clipping  = 0.05, which is asymmetric loss with different focusing factors for negative and positive instances and a probability shift of 0.05.
\textbf{We increased the number of epochs to 300 to allow each loss function to converge sufficiently and chose the best-performing loss function, binary cross-entropy, as the loss function for future experiments}.

\subsection{Hyperparameter Analysis}
\label{Sec: hyperparameter}
\begin{table}[!h]
    \centering
    \resizebox{0.75\linewidth}{!}{
    \begin{tabular}{c|c|c}
    \hline
     \textbf{Name} & \textbf{Initialization} & \textbf{Range for Tunning} \\
    \hline
    \centering  \multirow{3}{*}{Layer Structure} & \multirow{3}{*}{[768, 2048, 18]} & [768, 2048, 18], \\
    & & [768, 2048, 512, 18], \\
    & &  [768, 2048, 1024, 512, 18] \\
    \hline
    \centering Training Batch Size & 25000 & 25000, 10000, 5000\\
    \hline
    \centering Learning Rate & 0.001 & 0.001, 0.01, 0.1\\
    \hline
    \centering Epoch Number & 200 & 100, 200, 300, 400 \\
    \hline
    \centering Activation Function & GeLU & GeLU, ReLU, Leaky ReLU\\
    \hline
    \centering Dropout Rate & 0.5 &  0.2, 0.5, 0.6, 0.8\\
    \hline
    \end{tabular}}
    \caption{Hyperparameter Analysis.}
    \label{tab: hyperparameter analysis}
\end{table}
The greedy approach illustrates the pipeline of model tunning in Tab.(\ref{tab: hyperparameter analysis}). We first designed the
layer structure of the classification head. After that, we simultaneously designed the training batch size and learning rate, followed
by the choice of epoch numbers. Then is the selection of activations. Finally, we test the effect of different dropout rates to avoid overfitting.

\subsubsection{Layer Structure}
\begin{table}[!h]
    \centering
    \scriptsize
    \resizebox{0.55\linewidth}{!}{
    \begin{tabular}{c|c}
        \hline
	\multicolumn{1}{c|}{\textbf{Layer Structure}} & 
		\multicolumn{1}{c}{\textbf{Validation $F_1$ score (\%)}} \\
		\hline
         3 layers & 86.295 $\pm$ 0.000 \\
         4 layers & \textbf{86.526 $\pm$ 0.000} \\
         5 layers & 86.258 $\pm$ 0.000  \\
        \hline
    \end{tabular}}
    \caption{Comparison of the validation $F_1$ score of different layer structures. We use \textbf{bold} to represent the best result.}
    \label{tab: layer}
\end{table}
We test how different layer structures of the classification head (i.e., MLP) could affect the final result. Specifically, we use a 3-layer MLP with the layer structure of [768, 2048, 18], a 4-layer MLP with the layer structure of [768, 2048, 512, 18], and a 5-layer MLP with the layer structure of [768, 2048, 1024, 512, 18]. We report the averaged $F_1$ score with the standard deviation in Tab.(\ref{tab: layer}). Although different layer structures have little effect on the final performance, the 4-layer MLP outperforms others with a 0 variance. Also, it is reasonable to assume that increasing or decreasing the layer numbers will not boost the performance further, as the 4-layer MLP is better than both the 3-layer MLP and the 5-layer MLP. \textbf{Therefore, we decided to use the 4-layer MLP as the classification head for future experiments.} 

\subsubsection{Batch Size \& Learning Rate}
\begin{table}[!h]
    \centering
    \scriptsize
    \resizebox{0.7\linewidth}{!}{
    \begin{tabular}{c|c}
        \hline
	\multicolumn{1}{c|}{\textbf{Batch Size \& Learning Rate}} & 
		\multicolumn{1}{c}{\textbf{Validation $F_1$ score (\%)}} \\
		\hline
         25000, 0.001 & \textbf{86.531 $\pm$ 0.001} \\
         25000, 0.01 & 80.136 $\pm$ 0.003 \\
         25000, 0.1 & 59.163 $\pm$ 0.001  \\
         10000, 0.001 & 85.890 $\pm$ 0.000 \\
         10000, 0.01 & 80.292 $\pm$ 0.002 \\
         10000, 0.1 & 59.163 $\pm$ 0.000 \\
         5000, 0.001 & 84.721 $\pm$ 0.003 \\
         5000, 0.01 & 80.160 $\pm$ 0.002 \\
         5000, 0.1 & 0.000 $\pm$ 0.000 \\
        \hline
    \end{tabular}}
    \caption{Comparison of the validation $F_1$ score of different combinations of batch size and learning rate. We use \textbf{bold} to represent the best result.}
    \label{tab: batch_size}
\end{table}

\begin{figure}[h!]
  \centering
  \includegraphics[width=0.7\linewidth]{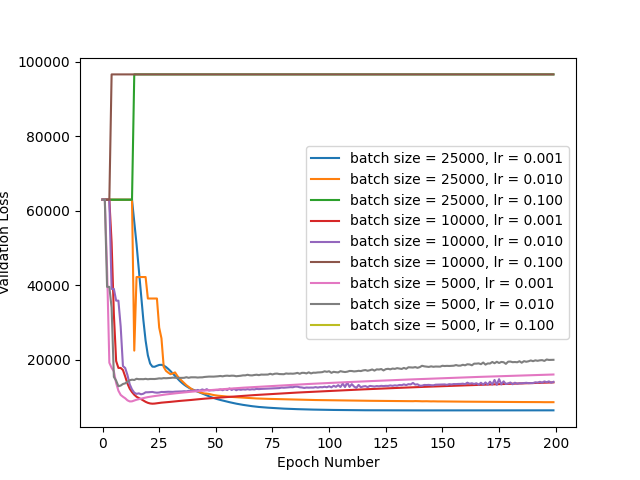}
  \caption{Validation loss with different batch sizes and learning rates.}
  \label{fig: batch_loss}
\end{figure}

The batch size and learning rate significantly determine the efficacy of gradient descent-based optimizers. A small learning rate may prolong the convergence of the training process, while a large one can accelerate it but at the risk of causing the loss function to diverge. In contrast, the batch size dictates the volume of data fed into the system. Notably, there is a correlation between the learning rate and batch size. Consequently, finding the optimal combination of batch size and learning rate through experimentation could potentially improve model performance.

There are three candidate values for both batch size and learning rate: [25000, 10000, 5000] and [0.001, 0.01, 0.1], respectively, creating nine distinct combinations. The mean $F_1$ score, along with the standard deviation, is reported in Tab.(\ref{tab: batch_size}), and the validation loss for each combination is visually depicted in Fig.(\ref{fig: batch_loss}). From the perspective of validation loss, most combinations demonstrate a consistent decrease during the training. However, it is observed that with a relatively high learning rate (e.g., 0.1), the training process can abruptly diverge, leading to an undefined validation $F_1$ score (i.e., when the batch size equals 5000 and the learning rate is 0.1, the validation $F_1$ score becomes 0). We also discovered that a large batch size can lead to a stable decrease in validation loss. Among all the combinations, the model performs best when the batch size = 25,000 and the learning rate = 0.001. \textbf{As a result, we decided to use the best-performing combination in the future.}

\subsubsection{Epoch Number}
\begin{table}[!h]
    \centering
    \scriptsize
    \resizebox{0.55\linewidth}{!}{
    \begin{tabular}{c|c}
        \hline
	\multicolumn{1}{c|}{\textbf{Epoch Number}} & 
		\multicolumn{1}{c}{\textbf{Validation $F_1$ score (\%)}} \\
		\hline
         100 & 84.586 $\pm$ 0.000 \\
         200 & 86.496 $\pm$ 0.000 \\
         300 & \textbf{86.554 $\pm$ 0.001}  \\
         400 & 85.770 $\pm$ 0.000 \\
        \hline
    \end{tabular}}
    \caption{Comparison of the validation $F_1$ score of different training epochs. We use \textbf{bold} to represent the best result.}
    \label{tab: epoch}
\end{table}
We are also curious about how different epoch numbers affect the model's performance. Therefore, we test the validation $F_1$ score under 4 different epoch numbers ranging from 100 to 400 (see Tab.(\ref{tab: epoch})). When the epoch size is 100, the validation $F_1$ is the lowest. This is probably due to the underfitting of the model, as the learning process may not fully converge. When we increase the epoch number to 300, the validation performance is further improved by approximately 2\%. If we further increase the epoch number, an overfitting problem occurs as the validation $F_1$ score drops. \textbf{As a result, 300 might be a suitable choice for the epoch number,} where the algorithm can fully converge without raising an overfitting issue.

\subsubsection{Activation Functions}
\begin{table}[!h]
    \centering
    \scriptsize
    \resizebox{0.6\linewidth}{!}{
    \begin{tabular}{c|c}
        \hline
	\multicolumn{1}{c|}{\textbf{Activation Function}} & 
		\multicolumn{1}{c}{\textbf{Validation $F_1$ score (\%)}} \\
		\hline
         GeLU & \textbf{86.554 $\pm$ 0.001} \\
         ReLU &  86.050 $\pm$ 0.000 \\
         Leaky ReLU & 86.267 $\pm$ 0.000 \\
        \hline
    \end{tabular}}
    \caption{Comparison of the validation $F_1$ score of different activation functions. We use \textbf{bold} to represent the best result.}
    \label{tab: activation}
\end{table}
Activation functions are important in the design of deep neural networks. We investigate the effect of activation functions on the classification head by testing the model performance under GeLU, ReLU, and Leaky ReLU, respectively. These three activation functions have some fundamental differences. Specifically, ReLU is the simplest and most widely used activation function. While it's computationally efficient and helps with the vanishing gradient problem, it suffers from a problem known as "dying ReLU," where neurons can sometimes get stuck in the state where they output zero for all inputs. Leaky ReLU is a variant of ReLU designed to fix the "dying ReLU" problem. Instead of outputting zero for negative inputs, it assigns a small negative slope. This way, the neuron can still learn and adjust its weights during backpropagation, even if the initial output for a given input is negative. GeLU is more complex than the previous two. It is a smooth, differentiable function approximating a Gaussian distribution's cumulative distribution function. GeLU has been shown to perform better than ReLU in some cases, but it is more computationally expensive. From our experiment results (see Tab.(\ref{tab: activation})), GeLU outperforms the others, but the gap between GeLU and Leaky ReLU is small. \textbf{Therefore, we may try both GeLU and Leaky ReLU in future experiments to see which one performs better.}

\subsubsection{Dropout Rate}
The dropout rate can effectively control the model’s ability to generalize. If the dropout rate is too low, the neural network might have too many parameters, which could cause overfitting. In this experiment section, we are interested in determining the dropout rate parameter for our development data to perform best and adapt the same dropout rate in the following experiments and the best model. We have four candidate values for the dropout rate: [0.2, 0.5, 0.6, 0.8], representing the probability of the node being dropped after activation. 

\begin{figure}[h!]
  \centering
  \includegraphics[width=0.7\linewidth]{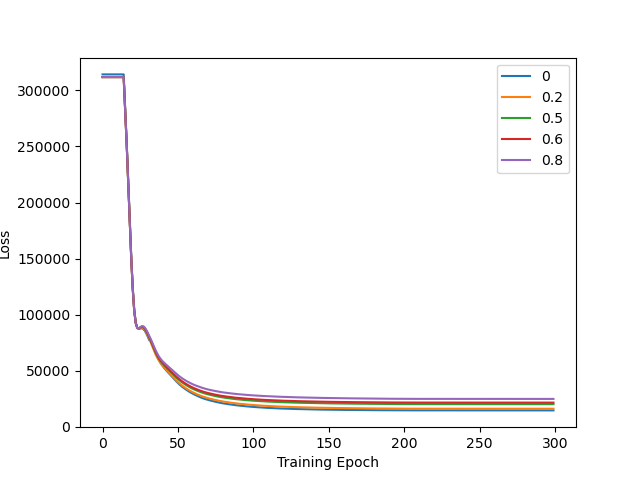}
  \caption{Training loss with different dropout rates on GeLU activation.}
  \label{fig: dropout loss}
\end{figure}
From the perspective of training loss, as shown in Fig.(\ref{fig: dropout loss}), all dropout rate values decreased training loss with more epochs, each resulting in about 50000 in total training loss after 50 epochs.  We then evaluate the model on development data using the $F_1$ score to investigate which dropout rate parameter will produce the best performance while also experimenting on two different activation functions from the previous experiments: GeLU and Leaky ReLU. 

\begin{table}[!h]
    \centering
    \scriptsize
    \resizebox{0.55\linewidth}{!}{
    \begin{tabular}{c|c}
        \hline
	\multicolumn{1}{c|}{\textbf{Dropout Rate}} & 
		\multicolumn{1}{c}{\textbf{Validation $F_1$ score (\%)}} \\
		\hline
          0.2, GeLU& 86.297 $\pm$ 0.001 \\
          0.5, GeLU & 86.554 $\pm$ 0.000 \\
          0.6, GeLU & \textbf{86.606 $\pm$ 0.000}  \\
          0.8, GeLU & 86.202 $\pm$ 0.000  \\
          0.2, Leaky ReLU & 85.927 $\pm$ 0.000 \\
          0.5, Leaky ReLU& 86.267  $\pm$ 0.001 \\
          0.6, Leaky ReLU& 86.349 $\pm$ 0.000 \\
          0.8, Leaky ReLU & 86.516 $\pm$ 0.001 \\
        \hline
    \end{tabular}}
    \caption{Comparison of the validation $F_1$ score of different dropout rate. We use \textbf{bold} to represent the best result.}
    \label{tab: dropout}
\end{table}

After comparing the model performance on development data from Tab.(\ref{tab: dropout}), we noticed the highest validation performance when the dropout rate = 0.6 and using the GeLU activation function. \textbf{Therefore, in all subsequent
experiments and the best model architecture, we set the dropout rate to 0.6 and use GeLU as the activation function.}

\subsection{Ablation Study}
\subsubsection{Dropout Rate}
We can see in the dropout experiments that the difference between keeping all nodes (i.e., dropout rate  = 0) and dropping out a portion of nodes (i.e., dropout rate  = 0.6) is considerable, with the dropout rate set to 0.6, and the validation $F_1$ score (\%) is 86.606. When the dropout rate is set to 0, the validation $F_1$ score (\%) is 85.955, resulting in a 0.65 difference (see Tab.(\ref{tab: dropout abl})). This effect is even more prominent when using the Leaky ReLU activation function.

This shows dropout does help with the overfitting problem. By leaving out a portion of the nodes in a layer, the model is less likely to rely on specific inputs and is forced to learn more robust features, increasing its generalization ability. \textbf{With this observation, dropout is kept in our best model, set to the best value found in parameter analysis with dropout rate  = 0.6.}
\begin{table}[!h]
    \centering
    \scriptsize
    \resizebox{0.55\linewidth}{!}{
    \begin{tabular}{c|c}
        \hline
	\multicolumn{1}{c|}{\textbf{Dropout Rate}} & 
		\multicolumn{1}{c}{\textbf{Validation $F_1$ score (\%)}} \\
		\hline
          0, GeLU& 85.955 $\pm$ 0.001 \\
          0.6, GeLU & \textbf{86.606 $\pm$ 0.000} \\
          0, Leaky ReLU & 85.269 $\pm$ 0.000 \\
          0.6, Leaky ReLU& 86.349 $\pm$ 0.001 \\
        \hline
    \end{tabular}}
    \caption{Comparison of the validation $F_1$ score of dropout and no dropout. We use \textbf{bold} to represent the best result.}
    \label{tab: dropout abl}
\end{table}

\subsection{Best Pipeline Results and Justification}

\BE{Best Pipeline}
To summarise the experimental result, the model obtains the best performance by leveraging 4 layers of MLP with GeLU activation function and 0.6 dropout rate to prevent overfitting. Moreover, the sum fusion is trained with the BCE loss. The implementation details of the hyperparameters are summarised in Tab.(\ref{tab:param}). The model is trained for 300 epochs with a learning rate of 1e-4 and batch size of 30000 (i.e., train with no validation data). The performance of the best pipeline achieves a 90.114\% public $F_1$ score on the Kaggle competition leaderboard. Since we only fine-tune with the classification head, our model is very lightweight; it could be trained within 4 minutes, and the model size is less than 25 MB. The device and software details are summarised in Tab.(\ref{table:components_settings}).

\BE{Justification}
Previous studies have shown that gMLP performs better than standard MLP. However, our experiments found that the extra gMLP layer will lead to the overfitting problem after 125 training epochs. This result infers that the feature extracted by the CLIP had already learned a good representation. The more complicated classification head may lead to a performance drop. Therefore, as for fine-tuning downstream tasks, the model complexity should be controlled. Moreover, we explored different fusion methods. Although the sum fusion by training two extra MLPs breaks the end-to-end training framework, it achieves a better result. Therefore, the conclusion is that fusion by a learning model is necessary, but the fusion part could not be complex.

\begin{table}[!h]
    \centering
    \resizebox{0.45 \linewidth}{!}{
    \begin{tabular}{c|c}
    \hline
     \textbf{Parameters} & \textbf{Value} \\
    \hline
    \centering Classification Head & MLP \\
    \hline
    \centering Activation Function & GeLU \\
    \hline
    \centering Dropout Rate & 0.6 \\
    \hline
    \centering Fusion Method &  Sum Fusion\\
    \hline
    \centering Loss Function & BCE\\
    \hline
    \centering Epochs & 300 \\
    \hline
    \centering Learning Rate & 0.001 \\
    \hline
    \centering Batch Size & 30000  \\
    \hline
    \end{tabular}}
    \caption{Parameters details of the best pipeline.}
    \label{tab:param}
\end{table}

\section{Conclusions}
In conclusion, we implemented the state-of-the-art (SOTA) techniques by utilising CLIP trained on a large dataset as the feature extractor for image and text modalities. Then, we fine-tuned the classification head and explored different classification heads, such as MLP and gMLP, with the composition of various feature fusion methods and loss functions. Finally, we obtained the model with the best performance of 90.114\% $F_1$ score in the Kaggele completion leaderboard due to the impressive transferability of CLIP and the fusion method. It is well worth noting that our approach is lightweight for both time and space complexity. The model could be trained in RTX3080 GPU in 4 minutes. The total model size is no more than 25 MB.

Although our model achieves the top 5 performance in the Kaggle leaderboard, some limitations remain. The first improvement could be the development of a unified learning architecture. The current training approach is not in the end-to-end schema, which results in a more complex hyperparameter tuning and deployment for real-world applications. Moreover, two-step learning can not be trained parallelly. The second training stage requires the output for the first step. Therefore, a possible improvement could be designing a model with two input sources simultaneously into a unified structure. However, due to the time limitation of the Kaggle completion, more experiments for the new model structures are required.

Moreover, we find the distribution of the data set is skewed. Further data preprocessing may lead to higher performance. To be specific, the majority of the distribution of label class is class 1. Therefore, it results in a shortcut solution for the model to predict class 1 with a low confidence level. The possible solution for addressing this is oversampling with the fewer classes or downsampling with the majority classes. Since our model consumes low training time, it is possible to train multiple models with different subsets of training data and ensemble the models for a more robust prediction. Finally, the shifting threshold may also be helpful by reducing the threshold value for the fewer classes. However, these methods introduce extra hyperparameters and training times, which is a trade-off between the performance and the computational cost.

Overall, the development of multimodal learning provides a solution to deal with the natural world by learning the comprehensive feature dependency across different data sources. The conventional research directions could be summarised as studying different interactions between different modalities, such as calculating the cosine similarity and consistency \cite{radford2021learning}, or using complex fusion method \cite{Kim2021ViLTVT}. Therefore, there are a few suggestions and insights:

\BE{Dynamic Fusion Methods} Traditional fusion methods, such as decision fusion and feature fusion, often lack flexibility. Recent advancements suggest that dynamic fusion methods, which allow the model to combine different modalities based on the input data adaptively, could provide more accurate and robust results \cite{tsai2019multimodal}.

\BE{Domain Adaptation and Transfer Learning} With the increasing complexity of multimodal tasks, leveraging knowledge from pre-existing models or similar domains is vital. Domain adaptation and transfer learning could help reduce the demand for labelled data and improve the models' generalisation ability \cite{tan2018survey}.

\BE{Robustness} Multimodal models should be robust to the absence or degradation of one or more modalities. This requires further research into methods for improving the resilience of such models to missing or noisy data.

\BE{Scalability and Efficiency} As the data volume and model complexity increase, the demand for computational resources also increases. Hence, it is essential to explore methods that can improve the scalability and efficiency of multimodal learning, such as lightweight model architectures and efficient training algorithms.

\clearpage
\bibliography{bib}
\bibliographystyle{ieeetr}
\end{document}